%% file: main.tex
\documentclass[sigconf, nonacm]{acmart}
\AtBeginDocument{%
  }

\setcopyright{none}
\renewcommand\footnotetextcopyrightpermission[1]{}
\copyrightyear{2018}
\acmYear{2018}
\acmDOI{XXXXXXX.XXXXXXX}
\acmConference[Conference acronym 'XX]{Make sure to enter the correct
  conference title from your rights confirmation email}{June 03--05,
  2018}{Woodstock, NY}
\acmISBN{978-1-4503-XXXX-X/2018/06}

\usepackage{booktabs}
\usepackage{amsmath}
\usepackage{amsthm}
\usepackage{multirow}
\usepackage{balance}
\usepackage{enumitem}
\usepackage{colortbl}
\usepackage{tabularx}
\usepackage{environ}

\theoremstyle{plain}

\theoremstyle{definition}
\newtheorem{definition}{Definition}

\newcommand{\rhat}{\hat{r}}

\newcommand{\phatb}{\hat{p}_{\mathrm{b}}}
\newcommand{\phath}{\hat{p}_{\mathrm{h}}}
\newcommand{\Cset}{\mathcal{C}}
\newcommand{\Qset}{\mathcal{Q}}
\newcommand{\Nin}{\mathcal{N}^{-}}
\newcommand{\items}{\operatorname{Items}}
\newcommand{\rb}{\bot}
\newcommand{\yhat}{\hat{y}}

\newcommand{\ci}[3]{$#1^{\scriptscriptstyle #3}_{\scriptscriptstyle #2}$}

\definecolor{gainred}{RGB}{200,78,42}
\definecolor{lossblue}{RGB}{70,130,180}
\definecolor{rowtint}{RGB}{244,244,246}
\definecolor{headtint}{RGB}{218,222,230}
\newcommand{\up}[1]{\,{\scriptsize\textcolor{gainred}{$\uparrow$#1}}}
\newcommand{\dn}[1]{\,{\scriptsize\textcolor{lossblue}{$\downarrow$#1}}}
\newcommand{\rowhl}{\rowcolor{rowtint}}
\newcommand{\thead}{\rowcolor{headtint}}
\newsavebox{\filltabA}\newsavebox{\filltabB}
\NewEnviron{filltab}[2]{%
  \sbox{\filltabA}{\setlength{\tabcolsep}{0pt}\begin{tabular}{#2}\BODY\end{tabular}}%
  \sbox{\filltabB}{\setlength{\tabcolsep}{1pt}\begin{tabular}{#2}\BODY\end{tabular}}%
  \setlength{\tabcolsep}{\dimexpr\dimexpr#1-\wd\filltabA\relax*65536/\numexpr\wd\filltabB-\wd\filltabA\relax\relax}%
  \begin{tabular}{#2}\BODY\end{tabular}}
\newcommand{\tablestyle}{\setlength{\aboverulesep}{0pt}\setlength{\belowrulesep}{0pt}\renewcommand{\arraystretch}{1.08}}
\definecolor{costgray}{RGB}{105,105,115}
\newcommand{\cost}[1]{\textcolor{costgray}{#1\,{\scriptsize$\times$}}}

\begin{document}

\title{Locating and Closing Information Gaps in Multi-Agent Systems}

\author{Yaxin Gong$^{1,2}$, Chongming Gao$^{1}$, Gangyi Zhang$^{2}$, Chenxiao Fan$^{1}$, Leyang Shen$^{3}$, 
\\
Jiakai Wang$^{2}$, Dong Wang$^{2}$, Yang Liu$^{2}$, Haoyan Liu$^{1}$, Xinting Hu$^{1}$\\
$^{1}$ University of Science and Technology of China \\ 
$^{2}$ Qwen Business Unit of Alibaba \quad $^{3}$ National University of Singapore \\ 
\texttt{gyx2022@mail.ustc.edu.cn} \\
}

\renewcommand{\shortauthors}{Yaxin Gong et al.}

\begin{abstract}
Multi-step Web tasks often require several LLM agents to gather and combine
evidence. The communication graph of a fixed workflow restricts which agents
can read each other's outputs: evidence retrieved by one agent may never
reach the agent that needs it, leaving an \emph{information gap} and an
incomplete or incorrect answer. Adding a connection can deliver the missing
evidence, but can also lead the receiver to adopt an incorrect upstream
conclusion that displaces a correct answer;
the final output alone does not separate the two.
We propose \textbf{S}elective \textbf{P}robing and \textbf{L}ink
\textbf{I}nsertion for \textbf{C}losing information gaps in multi-agent
\textbf{E}xecution (\textbf{SPLICE}), which locates information gaps per query
and decides what to adopt after adding an edge. After the original workflow
runs, SPLICE uses the query and node outputs to locate a possible gap, adds
one directed edge from a source to a receiver to deliver the relevant
evidence, and re-executes the receiver and affected downstream nodes,
producing a candidate answer. The adoption decision reads both the final
answers of the two runs and the process changes the edit caused inside the
workflow: how far the receiver moved, where its new content came from, and
whether the change reached the final answer. A selector trained on these
process features chooses between adopting the candidate answer and retaining
the original. On five benchmarks, SPLICE raises the task score on Web tool use
by $8.0\%$ and the mean score on four reasoning and planning benchmarks by
$5.7\%$ relative to the static skeleton, at $1.25$ to $1.44\times$ the
skeleton's cost, and transfers zero-shot to four unseen backbones.
\end{abstract}



\maketitle

\input{sections/01-intro}
\input{sections/02-related}
\input{sections/03-problem}
\input{sections/04-theory}
\input{sections/05-experiments}

\input{sections/07-conclusion}

\balance
\bibliographystyle{ACM-Reference-Format}
\bibliography{refs}

\appendix
\input{sections/A-appendix}

\end{document}

%% file: sections/01-intro.tex
\section{Introduction}
\label{sec:intro}

\begin{figure}[t]
  \centering
  \includegraphics[width=\columnwidth]{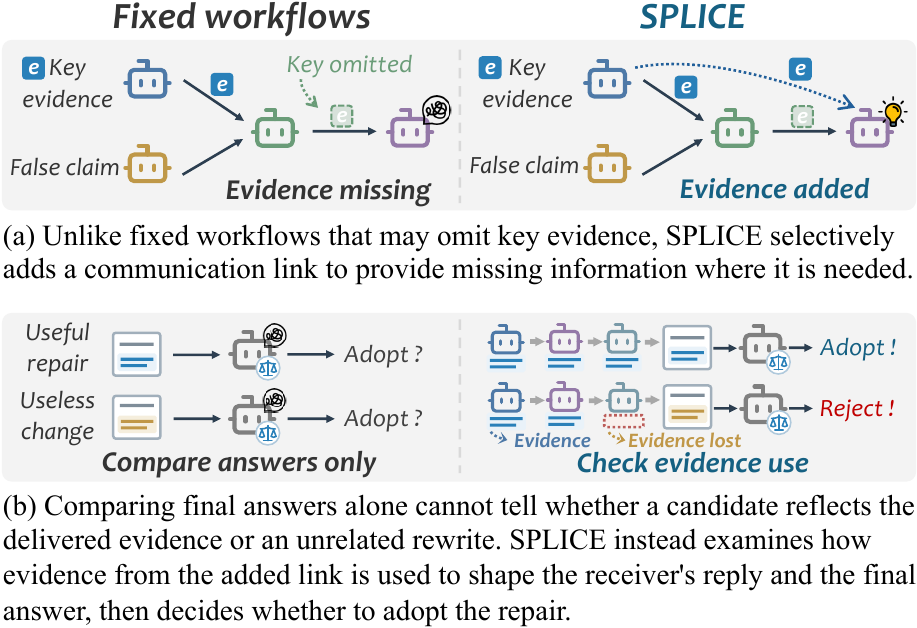}
  \caption{Comparison between existing fixed workflows (left) and our proposed SPLICE (right).}
  \label{fig:gap}
  \Description{Two stacked panels. The upper panel contrasts a workflow in
  which key upstream evidence is omitted, producing a false claim, with one
  in which an added edge delivers that evidence and the claim is corrected.
  The lower panel contrasts selecting between candidate outputs by answer
  text alone with selecting using the process evidence of each run.}
\end{figure}

A growing share of complex requests on the Web cannot be answered from any
single source, but require evidence gathered across multiple services,
interfaces, and pages\cite{trirec,agentbench}. Such multi-step information seeking is increasingly
carried out by pipelines of specialised LLM agents, each responsible for one
stage of retrieval or tool use~\cite{webwalker,selfrag}, which outperform single
agents on many tasks~\cite{macnet,gdesigner}. Together these stages form a
harness that gathers and analyses Web content. The pipeline's communication
graph determines which agents can read each other's outputs; the quality of
the final answer therefore depends on whether the evidence gathered across
these stages is grounded in the agent that has to act on it.

Consider a request to plan a cross-border trip. One agent looks up flights,
another checks visa rules, and a third assembles the itinerary. The flight
agent retrieves a route with a stop in a third country, but the visa agent
never receives this detail and applies the rules for a direct flight. The
final itinerary omits a required transit visa
(Figure~\ref{fig:gap}(a), left). In a
fixed workflow, an agent reads only the outputs of its direct
predecessors~\cite{moc}, even though the evidence it needs varies by query.
The relevant evidence is already available elsewhere in the workflow but
absent from the context of the agent that needs it. We call this an
\emph{information gap} (\S\ref{sec:prelim:rewire}).

A natural fix is to add a communication edge, deliver the evidence an
upstream agent holds to the receiver, and re-execute the affected downstream
nodes~\cite{moc,gdesigner,dats,daao} (Figure~\ref{fig:gap}(a), right). Adding
an edge, however, is not enough on its own. The inserted message can supply
the missing fact or lead the receiver to copy an incorrect upstream
conclusion and overwrite a correct
answer~\cite{upstreamoverwrite,wrongbutuseful}; the final output alone does
not separate the two (Figure~\ref{fig:gap}(b), left). The answer produced by
the edit and the re-execution---hereafter the \emph{candidate
answer}---therefore cannot be adopted automatically; whether to adopt the
candidate answer or retain the original must be decided after observing what
the edit did inside the workflow.

We draw inspiration from dendritic spine turnover: during learning, new
synaptic connections form, and subsequent local activity helps determine
which are stabilised or retracted~\cite{spineformation,spinestabilization}.
Following this ``form, then select by observed activity'' principle, this
paper proposes \textbf{S}elective \textbf{P}robing and \textbf{L}ink
\textbf{I}nsertion for \textbf{C}losing information gaps in multi-agent
\textbf{E}xecution (\textbf{SPLICE}). After the original workflow runs,
SPLICE uses the query and node outputs to locate an information gap, adds
one directed edge from a source to a receiver, sends the relevant evidence,
and re-executes the affected downstream nodes, producing a candidate answer.
The agents stay unchanged; the original answer is retained, and SPLICE
chooses between the candidate answer and the original by reading how outputs
changed between the original run and the trial.

SPLICE decides where to add an edge through two levels of diagnosis. At the
query level, a regressor predicts the base score each question would receive
under the static skeleton, and the trial budget goes to the questions with
the largest predicted room for improvement. At the edge level, SPLICE scores
each candidate edge by whether adding it will change the final output, and
adds the highest-scoring edge. The inserted message is not a relay of the
upstream answer; it grounds the Web evidence that the source agent holds but
the receiver's context lacks, closing the information gap.

After the edit, SPLICE chooses between the candidate answer and the
original. The two final answers do not separate several outcomes: the
inserted message is never used by the receiver, it is used but the revision
owes nothing to it, or the revision is absorbed downstream. SPLICE's
\emph{process-conditioned selection} therefore reads both the final answers
and the process changes the edit caused inside the workflow
(Figure~\ref{fig:gap}(b), right): how far the receiver moved, where its new
content came from (the message, the existing context, or neither), and
whether the change appears in the final answer. A selector trained on these
process features chooses between adopting the candidate answer and retaining
the original.

Our contributions are as follows.
\begin{itemize}[leftmargin=*]
\item We identify and characterise the \emph{information gap} in multi-agent
workflows, and propose SPLICE, a method that locates it per query, closes it
with one directed edge, and rolls back when verification is unfavourable.
\item We make the adoption decision from how outputs change inside the
workflow between the original run and the trial, which judges what to keep
more reliably than comparing final answers alone.
\item On five benchmarks SPLICE raises Web tool use by $8.0\%$ and the mean of
four reasoning benchmarks by $5.7\%$ relative to the static skeleton, composes
with the workflows of two published systems and a generic chain, and transfers
zero-shot to four backbones.
\end{itemize}

%% file: sections/02-related.tex
\section{Related Work}
\label{sec:related}

We review prior work along two lines that bear directly on this paper: how the
communication topology among agents is designed and adjusted~\cite{card}, and how, once a
modification produces a candidate answer, a choice is made among candidates.

\subsection{Multi-Agent Communication Topology}
\label{sec:related:topo}

Early multi-agent systems adopted a variety of fixed topologies for
inter-agent collaboration: CAMEL~\cite{camel} introduced communicative agents
that cooperate through role-playing, AutoGen~\cite{autogen} generalised this
paradigm to multi-agent dialogues, LLM-Debate~\cite{mad} leveraged a
multi-agent argumentation mechanism to correct hallucinations, and Generative
Agents~\cite{generativeagents} and AgentVerse~\cite{agentverse} facilitated
collaboration by simulating social environments or orchestrating expert
groups, with MetaGPT~\cite{metagpt} and ChatDev~\cite{chatdev} further
instantiating these fixed role divisions on software development pipelines.
As interaction patterns grew more complex, subsequent work began to model
agents explicitly as nodes in a graph and communication channels as edges,
optimising these graph structures to improve collaboration~\cite{eiblearner}:
GPTswarm~\cite{gptswarm}, DyLAN~\cite{dylan}, G-Designer~\cite{gdesigner}
and MACNET~\cite{macnet} each optimise the communication topology from a
different angle.
Building on this graph-based view, further work makes finer-grained edits to
the graph structure---MOC~\cite{moc} uniformly expands the range of upstream
outputs each agent can read, AgentPrune~\cite{agentprune} and
AgentDropout~\cite{agentdropout} remove edges judged redundant to improve
efficiency, and G-Safeguard~\cite{gsafeguard} intervenes topologically to
block malicious propagation.
More recently, DATS~\cite{dats}, AFlow~\cite{aflow},
MaAS~\cite{maas}, MASS~\cite{mass}, Codebook Agent~\cite{codebookagent}
and Nexa~\cite{nexa}
begin to decide the topology per query rather than committing one graph for
the entire dataset.
None of these methods, however, revisits the decision after observing what
the modified workflow produces: once an edge is added, its output is used
as the final answer. In this work we add an edge per query, run a trial,
and decide whether to adopt the candidate answer or retain the original
from how outputs change inside the workflow between the base run and the
trial.

\subsection{Selecting Among Candidate Outputs}
\label{sec:related:select}

When a modification produces a candidate answer, deciding whether to adopt it
is a selection problem.
The most basic approach places the original answer in a candidate pool and
picks the best: self-consistency~\cite{selfconsistency} takes a majority vote
over multiple samples, Self-Refine~\cite{selfrefine} and
Reflexion~\cite{reflexion} let the model reflect on and revise its own answer,
and multi-agent debate~\cite{mad} converges through multiple rounds
of discussion.
More refined selection mechanisms have since been proposed: Oracle
Gap~\cite{oraclegap} decomposes selector net gain into recoverable headroom,
verification signal quality, and harm to already-correct answers;
TrajSelector~\cite{trajselector} scores each reasoning step from the sampler's
own latent representations and aggregates those scores to pick a trajectory,
generative verifiers~\cite{genrm} score candidates at the token level,
MAS-ProVe~\cite{masprove} studies how well process verification actually works
across multi-agent frameworks, and Best-of-Evidence~\cite{bestofevidence}
keeps the candidate pool fixed while spending a limited budget on the evidence
actions that could change the choice; Consilience~\cite{consilience}
calibrates a conformal guarantee on each proposed communication action, and
DoVer~\cite{dover} verifies failure hypotheses through active intervention.
We compare against the last four of these decision rules on a shared
candidate pool in Table~\ref{tab:main-tvt}. TrajSelector reads the sampler's
latent representations, which the API backbones used here do not expose, so
it is not among the arms we run.
A related line of work addresses harmful messages. G-Safeguard~\cite{gsafeguard}
targets harm from an external adversary that constitutes a detectable
distribution shift; in normal collaboration, however, a message can be fluent
and well-reasoned yet wrong, and this kind of harm from honest agents has
never been managed as an intrinsic property of collaboration.
In all of the selection methods above, candidates are produced by independent
sampling or answer-level revision, and the original answer enters the choice
alongside the other candidates. Our candidates come from a per-query edge
addition and a trial; the original answer is the default returned when the
candidate is not adopted; and the adoption decision reads how outputs change
inside the workflow between the base run and the trial, not only the final
answers.

%% file: sections/03-problem.tex
\section{Preliminaries}
\label{sec:prelim}

This section sets up the notation used throughout the paper. We cast a
multi-agent workflow as a directed graph and then define candidate edges,
information gaps, and trials in turn.

\subsection{Multi-Agent Workflows as Graphs}
\label{sec:prelim:graph}

We adopt the graph formulation of multi-agent workflows used in prior
work~\cite{agentprune,gdesigner,moc}. A workflow is a directed acyclic graph
$G = (V, E)$ in which each node $v_i \in V$ is an LLM agent assigned a
subtask $x_i$, and each edge $(v_i, v_j) \in E$ means that the output of $v_i$
is readable by $v_j$. Given a query $q$ from a query set $\Qset$, nodes
execute in topological order: node $v_j$ produces an answer $a_j$ conditioned
on $q$, on its own subtask $x_j$, and on the outputs of its in-neighbours
$\Nin(v_j) = \{v_i : (v_i, v_j) \in E\}$. A designated terminal node emits
the final output $\yhat_q$, whose task score we write
$r_q = \mathrm{Score}(\yhat_q, y_q^{\star}) \in [0, 1]$, where $y_q^{\star}$
is the reference answer. We call the collection of all node answers in one
execution its \emph{trace}, $\tau_q = (a_1, \dots, a_{|V|}, \yhat_q)$.

Under this formulation, what $v_j$ reads from upstream is exactly the
concatenation of the answers of the nodes in $\Nin(v_j)$, placed in its
prompt~\cite{moc}. Two properties are worth stating explicitly, because the
problem we study follows from them. First, the query $q$ is visible to every
node, while the intermediate products the nodes derive from it travel only
along $E$; the visibility we refer to throughout is therefore the visibility
of these inter-node products, which is fixed at design time and does not vary
with the query. Second, a node forwards its own answer $a_i$ rather than the
upstream material it read; a piece of evidence that an upstream node derives
therefore either reaches $v_j$ in compressed form, having been rewritten at
every hop along the way, or does not reach it at all when the two sit on
branches that the skeleton never connects.

\subsection{Communication Rewiring}
\label{sec:prelim:rewire}

\paragraph{Candidate edges.}
Index the nodes by a topological order of $G$, so that $i < j$ implies $v_j$
is not an ancestor of $v_i$. The \emph{candidate edge set}
\begin{equation}
\label{eq:cand}
\Cset = \big\{ (v_i \to v_j) \;:\; i < j, \; v_i \notin \Nin(v_j) \big\}
\end{equation}
collects the forward edges missing from the skeleton, each of which would hand
material from an upstream node $v_i$ (the \emph{source}) directly to $v_j$
(the \emph{receiver}) without creating a cycle. Two kinds of source qualify:
nodes that already reach $v_j$ indirectly, and nodes on a parallel branch
that reach it not at all. $\Cset$ is determined by the skeleton alone and is
the same for every query; what varies across queries is which of its edges is
worth opening.

\paragraph{Information gaps.}
The candidate set collects every pathway that could be opened, but only some
of its edges correspond to a piece of evidence that is actually missing. For
a query $q$ and a candidate edge $(v_i \to v_j) \in \Cset$, we say the pair
constitutes an \emph{information gap} on $q$ when the source answer $a_i$
carries evidence that $v_j$ needs to answer but that is absent from its
upstream input $\bigcup_{v_l \in \Nin(v_j)} a_l$ and that $v_j$ does not
recover from $q$ on its own. Since $q$ is visible to every node, the gap is
the evidence that the source derived and that the receiver can neither read
nor reproduce; the definition is structural, stating that this evidence
exists at the source and is missing at the receiver. The cross-border example
in \S\ref{sec:intro} belongs to the second kind of source in~\eqref{eq:cand},
where source and receiver sit on parallel branches.
Membership is defined with reference to the reference answer $y_q^{\star}$
and is unobservable at deployment, where \S\ref{sec:method:diag} supplies
the ranking proxy used in its place.

\paragraph{Trials.}
Opening a candidate $e = (v_i \to v_j)$ and sending a message $m$ along it
yields the rewired graph $G \oplus (e, m) = (V, E \cup \{e\})$, in which $v_j$
reads $m$ in addition to its original inputs. Only $v_j$ and its descendants
need to be re-executed, producing a trial trace $\tau_q(e, m)$, a candidate
output $\yhat_q(e, m)$ and its score $r_q(e, m)$.

\paragraph{Gain.}
The score change of a trial relative to the base run is
$g_q(e, m) = r_q(e, m) - r_q$. Because scores lie in $[0, 1]$,
\begin{equation}
\label{eq:bounds}
-\,r_q \;\le\; g_q(e, m) \;\le\; 1 - r_q .
\end{equation}
We call a modification \emph{beneficial} (B), \emph{neutral} (S) or
\emph{harmful} (W) according to whether $g > 0$, $g = 0$ or $g < 0$.

\paragraph{Rewiring policies and blind commit.}
A rewiring policy makes three choices on each query: whether to trial,
$t_q \in \{0, 1\}$; which edges to trial, $S_q \subseteq \Cset$, together with
their messages; and, after the trial, an adoption decision
$d_q \in S_q \cup \{\rb\}$, where $\rb$ denotes retaining the original
answer $\yhat_q$. The policy space of rewiring strategies introduced in
\S\ref{sec:intro} can now be formalised through two properties.

\begin{definition}[Selectivity, retractability, blind commit]
\label{def:blind}
A policy is \emph{selective} if $(t_q, S_q)$ varies with $q$, and
\emph{retractable} if $d_q$ is a function of the trial trace $\tau_q(e, m)$
that can take the value $\rb$. A policy is a \emph{blind commit} if its
decision does not depend on anything observed after the trial, which takes
one of two forms: a \emph{no-trial commit} with $t_q = 0$ that modifies the
graph and returns the modified output as the final answer (predicted
topologies such as G-Designer, Nexa and MaAS fall here), and a
\emph{post-trial commit} with $t_q = 1$ but $d_q \neq \rb$ for every
query, which adopts unconditionally after every trial.
\end{definition}

%% file: sections/04-theory.tex
\section{Method}
\label{sec:method}

\begin{figure*}[t]
  \centering
  \includegraphics[width=\textwidth]{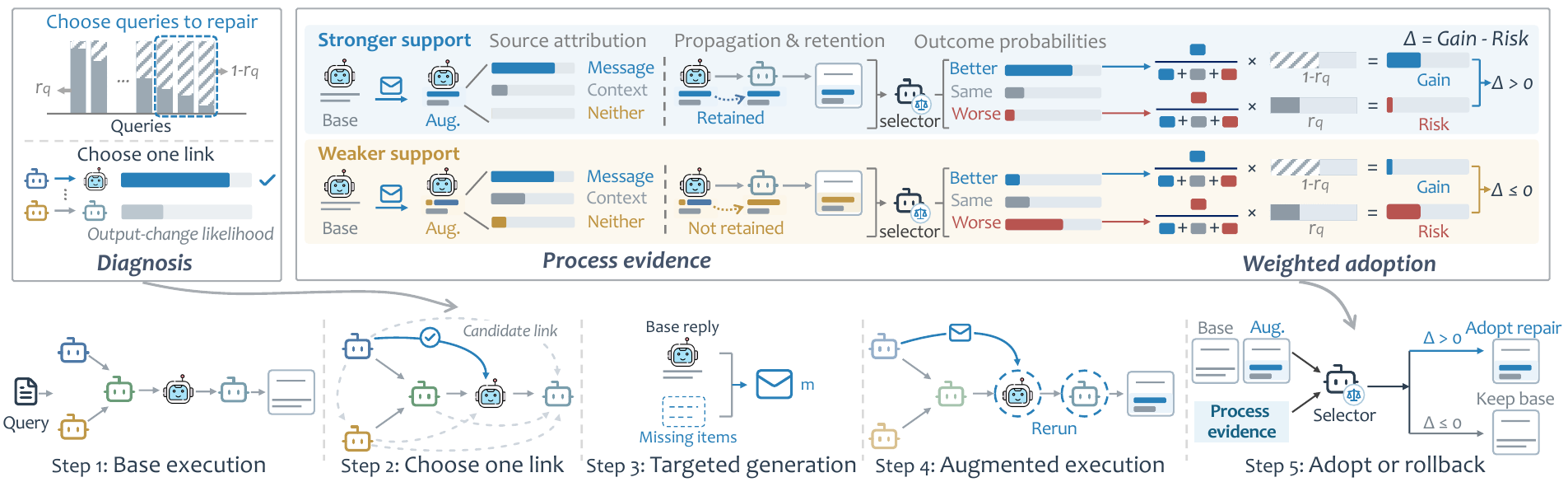}
  \caption{Overview of SPLICE. The four sub-decisions are made in one
  timeline: diagnosis selects which queries are worth a trial and which
  candidate edge to open, targeted generation writes the message that edge
  carries, and the trial is then executed. Only afterwards does verification
  read the resulting execution and decide whether to adopt the candidate
  answer or retain the original $\yhat_q$. The top half of the figure shows the core mechanism behind each step.}
  \label{fig:method}
  \Description{A left-to-right pipeline. A base run is diagnosed to locate a
  missing delivery, a message is generated for the selected candidate edge,
  the modified workflow is executed as a trial, and the process evidence of
  that trial drives an adopt-or-retain decision between the candidate
  answer and the original.}
\end{figure*}

SPLICE intervenes after the workflow has executed (Figure~\ref{fig:method}). For each selected query,
diagnosis picks one candidate edge and runs a single trial; at most
$B = b\,|\Qset|$ queries are trialled, where $b \in (0, 1]$ is the budget
fraction. The goal is to maximise the total score gain across all trialled
queries, counting retention of the original answer as zero gain. All
decisions before the trial rely only on the base trace $\tau_q$; only the
adoption decision may use the trial trace. This section is organised in two
parts: diagnosis and targeted generation decide \emph{where} to intervene
and \emph{what} to send; process-conditioned selection then decides, for
each trialled query, whether to adopt the candidate answer or retain the
original $\yhat_q$.

\subsection{Diagnosis and Targeted Generation}
\label{sec:method:diag}

\paragraph{Query-level diagnosis: whether to repair.}
By Eq.~\eqref{eq:bounds}, a trial can win at most the headroom $1 - r_q$,
which can be estimated before the trial. We fit the base score with a
regressor $\rhat_q = f_\omega(\psi(\tau_q))$ whose features $\psi$ read only
the base trace: whether the base output failed to parse, surface statistics of
the answer, and the agreement between the terminal node's answer and the final
output. None of these touches gold, and $f_\omega$ is fitted on the training
split only (\S\ref{sec:exp:setup}). We then set $t_q = 1$ for the $B$ queries
with the largest $1 - \rhat_q$.

\paragraph{Edge-level diagnosis: where to rewire.}
Within a selected query we pick one edge from $\Cset$ by ranking edges with a
gold-free estimate of the \emph{propagation probability}
\begin{equation}
\label{eq:rho}
\rho_{q,e} = P\big(\yhat_q(e, m) \neq \yhat_q \,\big|\, \tau_q, e\big),
\end{equation}
the probability that the trial changes the final output. It is estimated by a
light classifier $h_\eta$ whose features include the receiver's distance to
the terminal node, the agreement between the terminal answer and the final
output, and surface statistics of the receiver's answer. Its training label is
whether the final output changed, observable directly after a trial without
gold; $h_\eta$ is fitted on the training split only. At deployment every
feature of $h_\eta$ is computed from the base trace $\tau_q$ available
before the trial.

$S_q$ takes the single edge with the largest $\rho_{q,e}$.

\paragraph{Targeted generation: what to send.}
Instead of forwarding the source's answer $a_i$ verbatim, we have the source
organise its content against the receiver's gap. Let $\items(\cdot)$ be a
deterministic extractor returning the entities and numbers in a text; the gap
of receiver $v_j$,
\begin{equation}
\label{eq:gap}
M_{q,j} = \items(x_j) \setminus \bigcup_{v_l \in \Nin(v_j)} \items(a_l),
\end{equation}
contains the items its subtask mentions but none of its parents' outputs
cover. $M_{q,j}$ tells the generator which items to foreground. The message
is then
\begin{equation}
\label{eq:msg}
m = \mathrm{Gen}\big(a_i, M_{q,j}\big).
\end{equation}

\subsection{Process-Conditioned Selection}
\label{sec:method:s3}

An inserted message can be present in the receiver's input yet unused, can
be used without producing a revision that owes anything to it, or can move
the receiver only to have the change absorbed downstream. At the level of
the final answer all three can look alike, yet they mean very different
things. The adoption decision therefore reads the change the edit caused
inside the workflow between the base run and the trial.

\paragraph{Process evidence.}
After a trial, the base trace $\tau_q$ and the trial trace $\tau_q(e, m)$
form a pair, from which we read three groups of measurements. The first is
\emph{how far the receiver moved}: what its revision added, dropped, and how
large it was. The second is \emph{where the revision came from}: added
content attributed to what the message supplied, to what the receiver
already had, or to neither. The third is \emph{whether it survived}: how
far the change travelled toward the terminal node and whether the message's
content is still in the final answer. We write these as the \emph{process
evidence} $\phi_{q,e} = \phi(\tau_q, \tau_q(e, m))$; the full list of
fields is given in Appendix~\ref{app:evidence}.

\paragraph{Selector.}
We implement the adoption decision as a three-class language-model selector
$\pi_\theta(c \mid z_{q,e})$, $c \in \{\mathrm{B}, \mathrm{S}, \mathrm{W}\}$,
whose input
$z_{q,e} = (q, x_j, G, \rhat_q, \yhat_q, \yhat_q(e, m), \phi_{q,e})$
contains the query, the receiver's subtask, the workflow structure, the
predicted base score, the base and candidate outputs, and the process
evidence. Training labels $c^{\star} = \mathrm{sign}(g_q(e, m))$ are derived
from reference answers; no gold is used at inference. The selector is
initialised by supervised fine-tuning, which gives the reference policy
$\pi_{\mathrm{ref}}$, and then refined by direct preference
optimisation~\cite{dpo} on pairs $(c^{\star}, c^{-})$ with $c^{-} \neq c^{\star}$:
\begin{equation}
\label{eq:dpo}
\mathcal{L}(\theta) = -\,\mathbb{E}\Big[ \log \sigma\Big(
\beta \log \frac{\pi_\theta(c^{\star} \mid z)}{\pi_{\mathrm{ref}}(c^{\star} \mid z)}
- \beta \log \frac{\pi_\theta(c^{-} \mid z)}{\pi_{\mathrm{ref}}(c^{-} \mid z)}
\Big) \Big] .
\end{equation}

\begin{table*}[t]
\centering
\tablestyle
\caption{End-to-end results on five benchmarks. \textbf{Net} is the per-question fix$-$harm rate (\%). Best values per column are \textbf{bolded}.}
\label{tab:main-tvt}
\setlength{\tabcolsep}{3pt}
\newcommand{\mainsc}{1.1031}
\resizebox{\textwidth}{!}{%
\begin{filltab}{\mainsc\textwidth}{l cccc cccc ccc}
\toprule
\thead
& \multicolumn{4}{c}{\textbf{ToolBench} ($n{=}230$)}
& \multicolumn{4}{c}{\textbf{Four benchmarks} ($n{=}2{,}142$)}
& \multicolumn{3}{c}{\textbf{Total}} \\
\cmidrule(lr){2-5}\cmidrule(lr){6-9}\cmidrule(lr){10-12}
\thead
Method & Micro & Served & Net\,\% & Cost
& MuSiQue & DROP & 2Wiki & PlanCraft
& Mean & Net\,\% & Cost \\
\midrule
Static & 0.6347 & 0.3957 & -- & \cost{1.00}
& 0.4257 & 0.6734 & 0.7614 & 0.6988
& 0.6654 & -- & \cost{1.00} \\
\midrule
\rowhl
Always & 0.6514\up{.0167} & 0.3435\dn{.0522} & $+2.6$ & \cost{1.75}
& 0.4184\dn{.0073} & 0.7015\up{.0281} & 0.7649\up{.0035} & 0.6842\dn{.0146}
& 0.6725\up{.0071} & $+2.0$ & \cost{1.71} \\
Random & 0.6384\up{.0037} & 0.3870\dn{.0087} & $+0.9$ & \cost{1.67}
& 0.4343\up{.0086} & 0.7004\up{.0270} & 0.7607\dn{.0007} & 0.6491\dn{.0497}
& 0.6678\up{.0024} & $+1.9$ & \cost{1.71} \\
\rowhl
Self-cons. & 0.6282\dn{.0065} & 0.3609\dn{.0348} & $-1.7$ & \cost{3.00}
& 0.4346\up{.0089} & 0.6874\up{.0140} & 0.7627\up{.0013} & 0.6813\dn{.0175}
& 0.6692\up{.0038} & $-0.1$ & \cost{3.00} \\
Judge & 0.6624\up{.0277} & 0.4304\up{.0348} & $+6.1$ & \cost{2.67}
& 0.4418\up{.0161} & 0.6985\up{.0251} & 0.7717\up{.0103} & 0.6871\dn{.0117}
& 0.6781\up{.0127} & $+3.1$ & \cost{1.88} \\
\rowhl
Regressor & 0.6572\up{.0225} & 0.4304\up{.0348} & $+4.3$ & \cost{2.35}
& 0.4322\up{.0065} & 0.7011\up{.0277} & 0.7624\up{.0010} & 0.7281\up{.0292}
& 0.6808\up{.0154} & $+4.4$ & \cost{1.75} \\
\midrule
MOC & 0.6598\up{.0251} & 0.4304\up{.0348} & $+5.2$ & \cost{1.75}
& 0.4475\up{.0218} & 0.7218\up{.0484} & 0.7602\dn{.0012} & 0.6871\dn{.0117}
& 0.6831\up{.0177} & $+4.7$ & \cost{1.71} \\
\rowhl
Nexa & 0.6514\up{.0167} & 0.3435\dn{.0522} & $+2.6$ & \cost{1.09}
& 0.4314\up{.0057} & 0.6954\up{.0220} & 0.7643\up{.0029} & 0.7105\up{.0117}
& 0.6766\up{.0112} & $+2.5$ & \cost{1.42} \\
Nexa $+$ rb. & 0.6613\up{.0266} & 0.4130\up{.0174} & $+5.2$ & \cost{1.09}
& 0.4388\up{.0131} & 0.7006\up{.0272} & 0.7666\up{.0052} & 0.7193\up{.0205}
& 0.6818\up{.0164} & $+3.8$ & \cost{1.45} \\
\midrule
\rowhl
MAS-ProVe & 0.6580\up{.0233} & 0.4391\up{.0435} & $+4.3$ & \cost{2.71}
& 0.4363\up{.0106} & 0.7170\up{.0436} & 0.7569\dn{.0045} & 0.7485\up{.0497}
& 0.6883\up{.0229} & $+5.0$ & \cost{1.81} \\
Best-of-Evid. & 0.6419\up{.0072} & 0.4043\up{.0087} & $+1.7$ & \cost{1.85}
& 0.4276\up{.0019} & 0.7132\up{.0398} & 0.7634\up{.0020} & 0.7047\up{.0058}
& 0.6807\up{.0153} & $+3.1$ & \cost{1.41} \\
\rowhl
DoVer & 0.6419\up{.0072} & 0.4130\up{.0174} & $+0.9$ & \cost{1.33}
& 0.4384\up{.0127} & 0.6843\up{.0109} & 0.7689\up{.0075} & 0.6988
& 0.6737\up{.0083} & $+0.8$ & \cost{1.03} \\
Consilience & 0.6401\up{.0054} & 0.3957 & $+0.9$ & \cost{2.71}
& 0.4337\up{.0080} & 0.7125\up{.0391} & 0.7638\up{.0024} & 0.7368\up{.0380}
& 0.6868\up{.0214} & $+4.4$ & \cost{1.81} \\
\midrule
\rowhl
SPLICE ($b{=}0.3$) & 0.6769\up{.0422} & 0.4304\up{.0348} & $+7.0$ & \cost{1.27}
& 0.4575\up{.0318} & 0.6858\up{.0124} & 0.7633\up{.0019} & 0.7632\up{.0643}
& 0.6858\up{.0204} & $+4.7$ & \cost{1.11} \\
SPLICE ($b{=}0.5$) & \textbf{0.6856}\up{.0509} & \textbf{0.4478}\up{.0522} & $\mathbf{+8.7}$ & \cost{1.44}
& 0.4612\up{.0355} & 0.7158\up{.0424} & 0.7760\up{.0146} & 0.7778\up{.0789}
& 0.7031\up{.0377} & $+5.3$ & \cost{1.25} \\
\rowhl
\textbf{SPLICE (full)} & 0.6813\up{.0466} & 0.4391\up{.0435} & $+7.8$ & \cost{1.90}
& \textbf{0.4662}\up{.0405} & \textbf{0.7272}\up{.0538} & \textbf{0.7799}\up{.0185} & \textbf{0.7836}\up{.0848}
& \textbf{0.7101}\up{.0446} & $\mathbf{+7.5}$ & \cost{1.50} \\
\bottomrule
\end{filltab}}
\end{table*}

\paragraph{Weighted adoption.}
The selector scores three outcomes; the deployment action is binary, adopt
the candidate answer or retain the original $\yhat_q$. At deployment the
selector yields $\phatb = \pi_\theta(\mathrm{B} \mid z_{q,e})$ and
$\phath = \pi_\theta(\mathrm{W} \mid z_{q,e})$ for each trialled edge. The
adoption margin is
$\Delta_{q,e} = \phatb (1 - \rhat_q) - \phath\, \rhat_q$, weighting a
potential gain against the headroom $1 - \rhat_q$ and a potential loss
against the score $\rhat_q$ already in hand. The better the original
answer, the stronger the evidence needed to adopt the candidate. Neutral
outcomes carry no weight on either end. The decision is
\begin{equation}
\label{eq:adopt}
d_q =
\begin{cases}
\arg\max_{e \in S_q} \Delta_{q,e} & \text{if } \max_{e \in S_q} \Delta_{q,e} > 0, \\
\rb & \text{otherwise.}
\end{cases}
\end{equation}

%% file: sections/05-experiments.tex
\section{Experiments}
\label{sec:exp}

We organise our evaluation around four research questions.
\begin{itemize}[leftmargin=*]
\item \textbf{RQ1:} Does the full method beat the static skeleton and baselines?
\item \textbf{RQ2:} What does each component contribute?
\item \textbf{RQ3:} Does SPLICE locate the information gap it sets out to close?
\item \textbf{RQ4:} Does it generalise across host workflows, backbones and skeleton scales?
\end{itemize}

\subsection{Experimental Setup}
\label{sec:exp:setup}

\begin{table}[b]
\centering
\tablestyle
\setlength{\tabcolsep}{3pt}
\caption{Datasets and evaluation. Micro grades each required API by call and content; Served is the fraction of fully-scored requests; Token F1 and Plan success score the other four benchmarks against reference answers.}
\label{tab:datastats}
\resizebox{\columnwidth}{!}{%
\begin{tabular}{lllrr}
\toprule
\thead
Dataset & Task & Metric & Train & Test \\
\midrule
ToolBench & Web tool use & Micro (Served) & 535 & 230 \\
\rowhl
MuSiQue & Multi-hop QA & Token F1 & 1{,}680 & 360 \\
DROP & Discrete reasoning & Token F1 & 3{,}360 & 720 \\
\rowhl
2WikiMultihopQA & Multi-hop QA & Token F1 & 3{,}360 & 720 \\
PlanCraft & Embodied planning & Plan success & 1{,}596 & 342 \\
\bottomrule
\end{tabular}}
\end{table}

\paragraph{Datasets and evaluation.}
ToolBench~\cite{toolllm} is the primary workload (Table~\ref{tab:datastats}) and matches the Web-tool-use track adopted by broader agent-evaluation suites: agents issue tool calls through the
StableToolBench~\cite{stabletoolbench} virtual API server, which replays
archived responses so the same request yields the same result on any day,
and paired differences therefore reflect the communication repair rather
than service availability. PlanCraft is scored by executing the crafting
sequence against the benchmark's recipe library. Rollback is scoped to the
output layer and does not undo executed tool calls. Each dataset is split
at the question level; all learned components are fitted on the training
split only, and every reported number comes from a single evaluation on the
test split. The mean over the four reasoning and planning benchmarks is
weighted by test-set size, and the individual benchmark scores show how improvements vary across
question answering and planning. Appendix~\ref{app:data} gives skeleton structures and split details.

\paragraph{Baselines.}
We compare against five generic controls: Always, which adds all candidate
edges at once; Random, which opens one at random;
Self-consistency~\cite{selfconsistency}; a tuned LLM Judge; and a gain
Regressor. We also compare against the published systems
MOC~\cite{moc} and Nexa~\cite{nexa}, together with Nexa $+$ rollback, which
attaches our rollback step to the graph Nexa predicts, and against four
published selection
rules---MAS-ProVe~\cite{masprove},
Best-of-Evidence~\cite{bestofevidence}, Consilience~\cite{consilience} and
DoVer~\cite{dover}---all on the same skeletons, questions and splits, each
running its own procedure (details in Appendix~\ref{app:baselines}).

\paragraph{Implementation.}
All workflow nodes run on gpt-4o-mini; the adoption selector is a local
Qwen2.5-7B-Instruct~\cite{qwen25} trained with DPO. We report three budget
fractions $b \in \{0.3, 0.5, 1.0\}$. Intervals are $95\%$ question-paired
bootstrap CIs; cost is a dollar multiple over the static skeleton, with the
selector charged at its public hosted rate ($4.9\%$ of the total).
Appendices~\ref{app:selector} and~\ref{app:tables} give full details.

\subsection{End-to-End Performance (RQ1)}
\label{sec:exp:rq1}

The full method is ahead on both workloads. At $b{=}0.5$ it reaches
$0.6856$ Micro on ToolBench and a weighted mean of $0.7031$ on the four
benchmarks, $+0.0509$ and $+0.0377$ over the static skeleton, and its paired
difference against every competitor (Table~\ref{tab:paired-tvt}) stays above
zero. The half budget is a favourable operating point: it buys most of the
gain at $1.44\times$ the cost on ToolBench and $1.25\times$ on the four
benchmarks, trading a small amount of score for a lower cost than the full
budget. Opening edges without selection gives a different picture: Always and
Random fix about as many questions as they harm, a net rate of only
$+2.0\%$ and $+1.9\%$ on the four benchmarks, so the gain comes from where
to intervene.

The two calibers diverge in a way that is itself informative. Always and
Nexa raise Micro by $0.0167$ yet lose $0.0522$ on Served: opening an edge
unconditionally displaces answers that were already complete. The same
effect appears between our own budget settings. The half budget both serves
more requests ($0.4478$ against $0.4391$) and scores higher on Micro
($0.6856$ against $0.6813$) than the full budget, so the extra trials the
full budget buys crowd out some responses that needed no repair at all.

Attaching rollback to an existing method captures part of the gain but does
not close it: Nexa $+$ rollback reaches $0.6613$ on ToolBench, still $0.0243$
below the full method, and trails it by $0.0213$ on the four benchmarks. Targeted
diagnosis and process-conditioned selection add value beyond attaching
rollback to a predicted graph.

The gains hold under a different family of scoring rules from ToolBench's
Micro. MuSiQue, DROP and 2WikiMultihopQA are scored by standard token F1, and
PlanCraft by an executable verifier that steps through the recipe inventory,
and the weighted mean gain across them is $+0.0377$ at $b{=}0.5$. The same
method that helps on the Micro metric also helps under these separate rules,
so the improvement comes from the communication repair rather than from a fit
to one scoring signal.
On ToolBench, \S\ref{sec:exp:rq3loc} audits the same questions against
the API responses the workflow received.

\subsection{Component Contribution (RQ2)}
\label{sec:exp:rq2}

\begin{table}[t]
\caption{Query-level diagnosis calibration on ToolBench ($n{=}230$).
Bins are quintiles of predicted base score~$\rhat_q$.}
\label{tab:diag-cal}
\centering
\tablestyle
\resizebox{\columnwidth}{!}{%
\begin{tabular}{lcccccc}
\toprule
\thead
& Q1 (low) & Q2 & Q3 & Q4 & Q5 (high) & All \\
\midrule
Predicted & 0.392 & 0.548 & 0.651 & 0.736 & 0.847 & 0.635 \\
\rowhl
Actual    & 0.388 & 0.542 & 0.648 & 0.740 & 0.856 & 0.635 \\
Trialled  & 100\% & 100\% & 50\%  & 0\%   & 0\%   & 50\%  \\
\bottomrule
\end{tabular}}
\end{table}

\begin{table}[t]
\centering
\tablestyle
\caption{Component ablation on ToolBench with the four locating layers varied in turn. $\Delta$ is relative to Full ($b{=}0.5$); the intervals are 95\% paired bootstrap CIs over questions.}
\label{tab:abl-tvt}
\setlength{\tabcolsep}{3pt}
\begin{filltab}{\columnwidth}{l cccc}
\toprule
\thead
Configuration & Score & $\Delta$ & Net\,\% & Cost \\
\midrule
Static skeleton & 0.6347 & -- & -- & \cost{1.00} \\
\midrule
\rowhl
w/o diagnosis & 0.6813 & \ci{-.0043}{-.0162}{+.0073} & +7.8 & \cost{1.90} \\
$\rhat$ = constant & 0.6718 & \ci{-.0138}{-.0296}{-.0054} & $+6.1$ & \cost{1.44} \\
\rowhl
$\rhat$ = parse failure & 0.6648 & \ci{-.0208}{-.0401}{-.0146} & $+3.5$ & \cost{1.44} \\
$\rhat$ = length & 0.6642 & \ci{-.0214}{-.0410}{-.0147} & $+3.5$ & \cost{1.44} \\
\rowhl
edge $=$ random & 0.6654 & \ci{-.0202}{-.0392}{-.0137} & $+4.3$ & \cost{1.44} \\
edge $=$ predicted gain & 0.6671 & \ci{-.0185}{-.0372}{-.0021} & $+4.3$ & \cost{1.44} \\
\midrule
\rowhl
w/ verbatim forward & 0.6732 & \ci{-.0124}{-.0285}{-.0050} & $+5.2$ & \cost{1.44} \\
w/ untargeted msg & 0.6749 & \ci{-.0107}{-.0258}{-.0038} & $+6.1$ & \cost{1.44} \\
\rowhl
w/ key fact removed & 0.6745 & \ci{-.0111}{-.0246}{-.0052} & $+5.2$ & \cost{1.44} \\
w/ length-matched filler & 0.6697 & \ci{-.0159}{-.0351}{-.0082} & $+5.2$ & \cost{1.44} \\
\rowhl
w/o message (re-run) & 0.6665 & \ci{-.0191}{-.0375}{-.0112} & $+4.3$ & \cost{1.44} \\
w/ self-revision (matched budget) & 0.6689 & \ci{-.0167}{-.0349}{-.0012} & $+4.3$ & \cost{1.44} \\
\midrule
\rowhl
w/o rollback & 0.6727 & \ci{-.0129}{-.0307}{-.0052} & $+6.1$ & \cost{1.44} \\
adopt by $\phatb-\phath$ & 0.6738 & \ci{-.0118}{-.0290}{-.0041} & $+6.1$ & \cost{1.44} \\
\rowhl
adopt by predicted gain & 0.6789 & \ci{-.0067}{-.0160}{-.0005} & $+7.0$ & \cost{1.44} \\
w/ benefit head only & 0.6634 & \ci{-.0222}{-.0427}{-.0157} & $+3.5$ & \cost{1.44} \\
\rowhl
w/o process evidence & 0.6709 & \ci{-.0147}{-.0334}{-.0071} & $+5.2$ & \cost{1.44} \\
w/ shuffled process & 0.6699 & \ci{-.0157}{-.0351}{-.0081} & $+5.2$ & \cost{1.44} \\
\rowhl
w/ raw-trace selector & 0.6772 & \ci{-.0084}{-.0221}{-.0019} & $+6.1$ & \cost{1.44} \\
w/o attribution & 0.6780 & \ci{-.0076}{-.0210}{-.0012} & $+6.1$ & \cost{1.44} \\
\rowhl
attribution only & 0.6760 & \ci{-.0096}{-.0238}{-.0028} & $+5.2$ & \cost{1.44} \\
w/ SFT selector & 0.6818 & \ci{-.0038}{-.0160}{+.0081} & $+7.8$ & \cost{1.44} \\
\midrule
\rowhl
\textbf{Full} ($b{=}0.5$) & \textbf{0.6856} & -- & $\mathbf{+8.7}$ & \cost{1.44} \\
\bottomrule
\end{filltab}
\end{table}

We verify each component along the four layers of locating.

\paragraph{Layer 1: Query-level locating.}
Table~\ref{tab:diag-cal} bins the $230$ test questions into quintiles by
predicted base score~$\rhat_q$ and reports the mean predicted score, the mean
actual score and the fraction trialled at budget $b{=}0.5$.
Predicted and actual scores track closely.
The budget concentrates entirely on the two lowest quintiles, which are
trialled at 100\%, while the two highest quintiles are skipped entirely.
Table~\ref{tab:abl-tvt} corroborates this: removing diagnosis altogether raises
cost without raising the score, and replacing the predicted score with a
constant, a parse-failure flag or answer length all fall significantly
below the full diagnosis.

\paragraph{Layer 2: Edge-level locating.}
On the same $535$ training queries, the edge selected by $\rho$, which is the
edge the full method uses, produces an output change on $115$ questions, of
which $70$ improve and $45$ worsen, a fix ratio of $0.61$. A randomly chosen
edge produces a change on $96$ questions, $44$ improving and $52$ worsening, a
fix ratio of $0.46$. The edge diagnosis thus selects edges that are not merely
more likely to change the output but more likely to change it for the better.
In the ablation table, random edge selection falls $-0.0202$ behind the full
method, and ranking edges by a predicted-gain head falls $-0.0185$ behind
it: $\rho$ selects better edges than a direct gain forecast, since the gain
depends on what the receiver does with the message and is only measurable
after the trial.

\paragraph{Layer 3: Message-level locating.}
The message ablations form a ladder that holds the edge and the re-execution
fixed and varies only what is sent. At the bottom, re-running the same nodes
with no edge and no message at the same $1.44\times$ budget scores $0.6665$.
Opening the edge with length-matched filler adds $0.0032$. Removing only the
key factual statement while keeping the rest of the message intact gives
$0.6745$, and the complete message gives $0.6856$. Of the $0.0159$ separating
filler from the full method, roughly seven tenths is carried by that single
factual statement, confirming that the message locates a specific piece of
missing evidence. On the same queries, $115$ answers change when a targeted
message is delivered against $68$ without one, so the adoption rule has
$69\%$ more candidates to choose among.
Spending the same $1.44\times$ budget on receiver self-revision with the same
selector, rather than on a targeted message, reaches $0.6689$, so at matched
cost the gain comes from delivering the located evidence rather than from the
extra computation alone.
Appendix~\ref{app:cases} reproduces two archived edges, one a ToolBench edge carrying an API payload downstream.

\paragraph{Layer 4: Process-evidence verification.}
Removing process evidence entirely costs $-0.0147$, and shuffling it costs
$-0.0157$, confirming that the features carry a real signal. The attribution
group isolates the most diagnostic layer: removing it alone costs $-0.0076$,
so knowing whether new content originated from the message or from the
receiver's own resampling has independent value for the adoption decision.
A selector reading the same base and trial traces as raw text, with the same
model, training data, and candidate pool but without the hand-computed
attribution features, reaches $0.6772$ and trails the full method by $-0.0084$,
so the explicit process-attribution features add adoption value on top of
access to the same trace.
Adopting by the unweighted difference $\hat p_b-\hat p_h$ falls behind the
asymmetric default by $-0.0118$ (CI clear of zero), showing that weighting a
candidate against the headroom on one side and the score already in hand on
the other is doing work.
Replacing the selector altogether with a predicted-gain regressor costs
$-0.0067$ (CI $[-.0160, -.0005]$), so the three-class process-conditioned
selector improves adoption beyond a direct gain forecast on top of the same
adoption rule. These results support combining process evidence with the predicted
base score when deciding whether to adopt a repair. The process
features help distinguish beneficial from harmful changes, while
the base score determines their relative weights: a higher predicted
score increases the penalty for potential harm and reduces the
weight assigned to potential improvement.

\subsection{Gap Localisation (RQ3)}
\label{sec:exp:rq3loc}

\begin{table}[t]
\centering
\tablestyle
\caption{Gap-hit precision@1 on the held-out ToolBench test set, with two judges supplying the gap labels and $\rho$ compared against three heuristic edge-selection rules.}
\label{tab:gaphit}
\setlength{\tabcolsep}{10pt}
\begin{tabular}{lcc}
\toprule
\thead
Edge selection rule & Precision@1 & $95\%$ CI \\
\midrule
Random reachable edge        & $0.41$ & $[.34,\,.48]$ \\
\rowhl
Terminal-distance (nearest)  & $0.49$ & $[.42,\,.56]$ \\
Content coverage             & $0.55$ & $[.48,\,.62]$ \\
\rowhl
$\rho$ (SPLICE)                & $\mathbf{0.67}$ & $[.60,\,.74]$ \\
\bottomrule
\end{tabular}
\end{table}

\paragraph{Pre-trial gap-hit precision.}
We first test whether the edge selector locates real gaps before any trial.
On the held-out test set, two judges label for each question which candidate
edges correspond to a true information gap, a source-held fact absent from the
receiver's upstream context and not recovered from $q$; we take this as ground
truth and measure how often each rule's top-ranked edge is a true gap
(Table~\ref{tab:gaphit}). Ranking by $\rho$ reaches a precision@1 of $0.67$,
above random reachable edges ($0.41$), the terminal-distance heuristic
($0.49$) and content coverage ($0.55$): the pre-trial selector favours real
gaps over edges that merely change the output.

\begin{table}[t]
\centering
\tablestyle
\caption{Gap-localisation audit on $n{=}120$ adopted repairs, independently scored by two judges across four conditions whose joint rate is reported in the bold row.}
\label{tab:gapaudit}
\resizebox{\columnwidth}{!}{%
\begin{tabular}{lcc}
\toprule
\thead
Condition & DeepSeek-v4-pro & Claude-opus-4-6 \\
\midrule
Source holds the fact        & $0.88$ [.82,.93] & $0.86$ [.79,.91] \\
\rowhl
Receiver lacked the fact     & $0.91$ [.85,.95] & $0.89$ [.83,.94] \\
Message delivered the fact   & $0.85$ [.78,.90] & $0.83$ [.76,.89] \\
\rowhl
Revision used the fact       & $0.82$ [.75,.88] & $0.80$ [.72,.86] \\
\midrule
\textbf{All four hold}       & $\mathbf{0.74}$ [.66,.81] & $\mathbf{0.71}$ [.63,.79] \\
\bottomrule
\end{tabular}}
\end{table}

\paragraph{Post-adoption gap audit.}
We then audit whether an adopted edge closes a real gap. Two judges unrelated
to the workflow backbone and the selector independently label a random sample
of $120$ adopted repairs against four conditions: the source holds the fact,
the receiver
lacked it, the message delivered it, and the revision used it
(Table~\ref{tab:gapaudit}). All four conditions hold on $0.74$ and $0.71$ of
the sample under the two judges, with substantial agreement (Cohen's
$\kappa{=}0.77$). The audit separates an output change from a closed gap: the
majority of adopted repairs carry a fact the source already held and the
receiver lacked, deliver it, and use it in the corrected answer.

Across the $28{,}560$ edges of the offline probe pool, $613$ ($2.1\%$) satisfy
all four conditions, split almost evenly between sources on a parallel branch
the skeleton never connects ($315$) and distant-ancestor sources whose content
was rewritten away hop by hop ($298$); both are gaps a fixed topology cannot
cover. Appendix~\ref{app:cases} reproduces one closed gap of each kind
verbatim.

\begin{table}[t]
\centering
\tablestyle
\setlength{\tabcolsep}{8pt}
\caption{Evidence support on the held-out ToolBench test set
($n{=}230$); the first three rows sum to one, and the judging protocol is
detailed in Appendix~\ref{app:grounding}.}
\label{tab:grounding}
\begin{tabular}{lcc}
\toprule
\thead
Measure & Static & SPLICE \\
\midrule
Fact-support rate       & $0.71$ [.65,.77] & $\mathbf{0.83}$ [.78,.88] \\
\rowhl
Partially-supported rate & $0.11$ [.07,.15] & $0.08$ [.05,.12] \\
Unsupported-claim rate  & $0.18$ [.13,.23] & $\mathbf{0.09}$ [.06,.13] \\
\rowhl
Task completion rate    & $0.63$ [.57,.70] & $\mathbf{0.74}$ [.68,.80] \\
\bottomrule
\end{tabular}
\end{table}

\paragraph{Evidence support.}
The audit so far reads the workflow trace; we close RQ3 at the answer level by
checking whether the content a closed gap introduces is itself grounded in an
API response. On the $230$ held-out test questions, two judges read the API
responses the workflow actually received and verify each concrete claim, a
number, name, date or URL, against them, labelling each as supported,
partially supported or unsupported (three rates summing to one;
Table~\ref{tab:grounding}). SPLICE
raises the fact-support rate from $0.71$ to $0.83$ and the task completion rate
from $0.63$ to $0.74$, and lowers the unsupported-claim rate from $0.18$ to
$0.09$. The gain therefore lands on claims that an API response backs rather
than on answer specificity alone, so the improvement reflects evidence
support rather than longer or more detailed answers.

\begin{table}[t]
\centering
\tablestyle
\setlength{\tabcolsep}{4pt}
\caption{A held-out ToolBench case where SPLICE closes an API-payload
gap: the UUID returned at node~0 reaches the terminal only after the added
edge delivers it. Full archived exchange in Appendix~\ref{app:cases}.}
\label{tab:uuid-case}
\begin{tabularx}{\columnwidth}{p{0.24\columnwidth}X}
\toprule
\thead
Checkpoint & Archived observation \\
\midrule
Node~0 return & API payload: $u^\ast$ + captcha PNG. \\
\rowhl
Terminal input & Parents (nodes~1--2): no payload and no $u^\ast$. \\
Repair & Open $0\!\to\!3$: deliver payload to terminal. \\
\rowhl
Answer & Static: ``no API results''; SPLICE: $u^\ast$ + image. \\
\bottomrule
\end{tabularx}
\end{table}

Table~\ref{tab:uuid-case} makes the answer-level audit concrete: the repaired answer reuses
the value returned at node~0, while the static terminal had no access to any
API result.

\subsection{Generalisation and Boundaries (RQ4)}
\label{sec:exp:rq4}

\paragraph{Composability.}

\begin{table}[t]
\centering
\tablestyle
\caption{Composability on three host workflows; $p$ compares SPLICE against each host's own static output.}
\label{tab:host}
\resizebox{\columnwidth}{!}{%
\begin{tabular}{llrrrr}
\toprule
\thead
Host & Bench & Static & Always & SPLICE & $p$ \\
\midrule
Chain       & ToolBench & 0.4222 & 0.4604\up{.0382} & 0.4832\up{.0610} & 0.0003 \\
\rowhl
AgentPrune~\cite{agentprune}  & ToolBench & 0.4180 & 0.4428\up{.0248} & 0.4713\up{.0533} & 0.0003 \\
G-Designer~\cite{gdesigner}  & ToolBench & 0.3995 & 0.4212\up{.0217} & 0.4433\up{.0438} & 0.0008 \\
\midrule
\rowhl
Chain       & 2Wiki & 0.6009 & 0.5942\dn{.0067} & 0.6242\up{.0233} & 0.0828 \\
AgentPrune  & 2Wiki & 0.5616 & 0.5728\up{.0112} & 0.6076\up{.0460} & 0.0276 \\
\rowhl
G-Designer  & 2Wiki & 0.6625 & 0.6689\up{.0064} & 0.6833\up{.0208} & 0.0100 \\
\bottomrule
\end{tabular}}
\end{table}

SPLICE adds value on top of existing workflows (Table~\ref{tab:host}). Across six
host-by-benchmark cells it improves on each host's own output: by $+0.044$ to
$+0.061$ on the three ToolBench hosts ($p \le 0.0008$) and by $+0.021$ to
$+0.046$ on the three 2Wiki hosts. The 2Wiki-Chain cell is the informative
one: its candidate pool is net negative, so opening every edge (Always) loses
$0.0067$, while SPLICE still gains $+0.0233$ because rollback keeps the harmful
edits out. Per-query diagnosis and rollback thus act as a layer on top of an
existing orchestration rather than a replacement for its topology.

\paragraph{Backbone, scale and budget.}
\begin{table}[t]
\centering
\tablestyle
\caption{Backbone transfer on DROP, 2Wiki and PlanCraft, with the selector fitted once on gpt-4o-mini and reused on four unseen backbones with no retraining.}
\label{tab:backbone-tvt}
\setlength{\tabcolsep}{4pt}
\resizebox{\columnwidth}{!}{%
\begin{tabular}{lrrrcr}
\toprule
\thead
Backbone & Static & Always & SPLICE & $95\%$ CI & $p$ \\
\midrule
Kimi-K2.6      & 0.7150 & 0.7436 & 0.7652\up{.0502} & [.026,\,.076] & $1.3\!\times\!10^{-4}$ \\
\rowhl
DeepSeek-V3.2~\cite{deepseekv32}  & 0.7245 & 0.7167 & 0.7758\up{.0513} & [.030,\,.076] & $1.2\!\times\!10^{-5}$ \\
Qwen3.7-Max    & 0.8355 & 0.8406 & 0.8673\up{.0317} & [.014,\,.052] & $1.2\!\times\!10^{-3}$ \\
\rowhl
Qwen3.6-Plus   & 0.7636 & 0.7714 & 0.7921\up{.0284} & [.012,\,.047] & $1.7\!\times\!10^{-3}$ \\
\bottomrule
\end{tabular}}
\end{table}

\begin{figure}[t]
  \centering
  \includegraphics[width=\columnwidth]{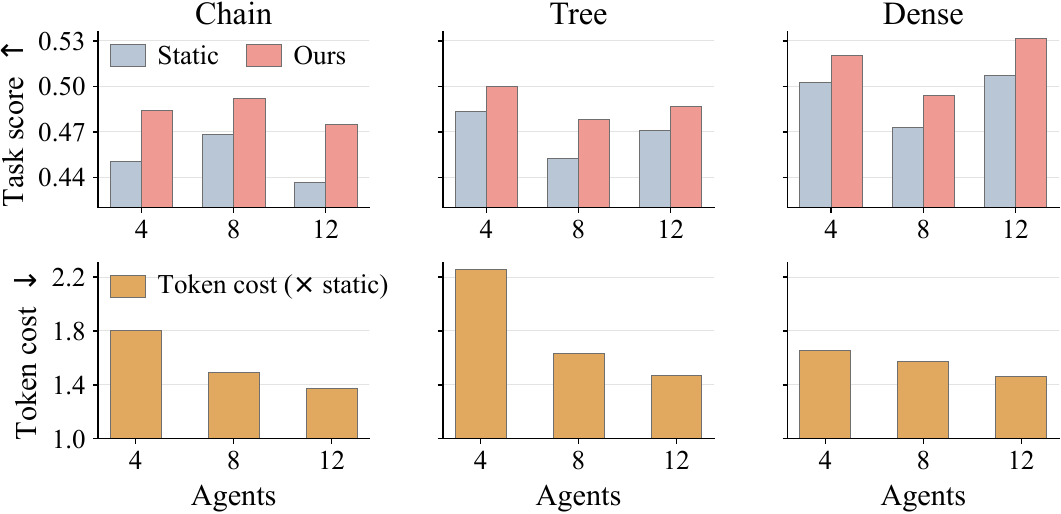}
  \caption{Skeleton scan on DROP across three topology families (Chain, Tree,
  Dense) and three agent scales (4, 8, 12).}
  \label{fig:skeleton}
  \Description{Two rows of panels: top row shows task score for Chain, Tree
  and Dense topologies at 4, 8 and 12 agents, with SPLICE consistently above
  Static; bottom row shows the corresponding cost multiples.}
\end{figure}

\begin{figure}[t]
  \centering
  \includegraphics[width=\columnwidth]{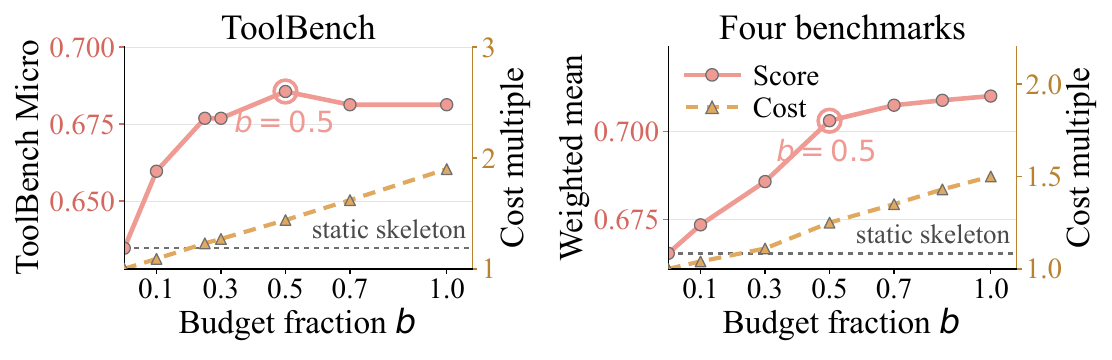}
  \caption{Score and cost against the trial-budget fraction $b$ on ToolBench and the four reasoning and planning benchmarks.}
  \label{fig:budget}
\end{figure}

The method holds as the setting changes. The selector is fitted once on
gpt-4o-mini and never retrained, yet transfers to four unseen backbones with
gains of $+0.028$ to $+0.051$ (Table~\ref{tab:backbone-tvt}, all
$p \le 2\times10^{-3}$), so the gain does not fade as the backbone grows
stronger. Across nine topologies and scales the direction is positive throughout (Figure~\ref{fig:skeleton}), as the number of candidate edges
$|\Cset|$ ranges from $2$ to $55$ and
the gains run from $+0.0154$ to $+0.0383$. The budget sweep separates the two workloads:
ToolBench Micro rises monotonically with $b$ up to a peak at $b{=}0.5$ and declines at $b{=}1.0$, while the four reasoning and planning benchmarks
rise and then saturate for $b\geq 0.5$ (Figure~\ref{fig:budget}), so the
operating point $b{=}0.5$ sits on the knee for both workloads at matched
cost.

The gain is positive across all nine topologies and scales, and its size
follows how much the terminal structure lets a local edit move.
The candidate count does not determine the gain monotonically: the smallest
pool ($|\Cset|{=}2$) and the largest one ($|\Cset|{=}55$) both improve, while
the nine gains interleave between $+0.0154$ and $+0.0383$ rather than rising
with $|\Cset|$. SPLICE therefore buys better gap localisation rather than a larger
candidate pool or additional trials per query. The size of each gain tracks
the room the terminal structure leaves for a single rewritten edge to move
the final answer, as the per-cell counts in Figure~\ref{fig:skeleton} show
across chain and tree families alike.

%% file: sections/07-conclusion.tex
\section{Conclusion and Future Work}
\label{sec:conclusion}

When multiple LLM agents collaborate on Web content, the quality of the
final answer depends on whether each cross-agent transfer of evidence
reaches the party that needs it.
We proposed SPLICE, which after the original workflow has executed locates
an information gap per query, closes it with one directed edge, re-executes
the affected downstream nodes, and chooses between the candidate answer and
the original from how outputs change inside the workflow between the base
run and the trial.
Across five benchmarks SPLICE raises the task score on Web tool use by
$8.0\%$ and the mean on four reasoning and planning benchmarks by $5.7\%$
relative to the static skeleton at $1.25$ to $1.44\times$ its cost, composes
with the workflows of two published systems and a generic chain, and
transfers zero-shot to four unseen backbones.
Intervention analysis and ablations together show that the final answer
alone cannot say what should be kept; the adoption decision has to read
the process trace an edit leaves inside the workflow between the original
run and the trial.

As Web agent pipelines scale to more sources and longer tool-use chains,
cross-agent evidence transfer becomes the reliability pivot of
collaborative systems. The present work operates with a single gap and one
real trial per query.
Pushing process evidence from agent replies down to pages, documents, and
entities, and extending the method from read-only Web to writing and
transactional APIs, are natural next steps: in these settings cross-source
corroboration can tighten gap identification, and single-edge repair can
grow into multi-step repair in which process features replace a real trial
as a cheaper adoption criterion.
We plan to pursue these directions toward Web multi-agent systems in which
diagnosis, repair, and verification run through the full collaboration
lifecycle, providing a more reliable foundation for increasingly complex
Web mining and information integration tasks.

%% file: sections/A-appendix.tex
\section{Reproducibility and Supplementary Tables}
\label{app:tables}

\subsection{Data and splits}
\label{app:data}
Table~\ref{tab:datastats} in the main text lists the datasets and splits.
Splits are taken over questions rather than edges so that the candidates of
a question never straddle a boundary.
ToolBench had no split field in the archive, so we added one with the same
$70/15/15$ ratio under a fixed seed and froze it before any evaluation; its
validation and test portions are then evaluated as a single held-out split
of $230$ requests, and every quantity that the other benchmarks select on
validation is selected within the ToolBench training split instead.

\paragraph{Plan execution on PlanCraft.}
The inventory is parsed from the item--count lines of the prompt, and the
crafting sequence is read from the workflow's final answer as a list of item
names. The checker executes that sequence one step at a time against the
benchmark's own recipe library: for each item it reads the recipe, counting
required quantities from the occurrences of each key in the pattern for
shaped recipes and from the ingredient list for shapeless ones, then
consumes the ingredients and adds the product when the inventory holds
enough. A plan scores $1$ when every step is executed this way and the
target item ends up in the inventory. For the infeasible instances a
reachability search over the same library establishes that no sequence
produces the target, and the workflow must report the task as impossible. A
final answer from which no crafting sequence can be parsed scores $0$, the
same convention the other benchmarks use for unparsable output.

ToolBench contributes all $765$ solvable requests in the six subsets of
StableToolBench. Both
skeletons are four-node workflows, three sub-task nodes and one aggregator,
and the candidate edges are the forward edges missing from the skeleton. Tool-enabled nodes first issue tool calls and then generate their
outputs from the returned API content.
Table~\ref{tab:adj} lists every skeleton as an adjacency list
(\textit{receiver}:\{\textit{sources}\}; nodes with no incoming edge
omitted), with the $|\Cset|$ column counting the candidate edges
Eq.~\eqref{eq:cand} admits on that skeleton. The scan families name the
generator rather than the shape of each instance: the parallel-spine
generator (\textit{Chain}) degenerates to a diamond at $N{=}4$ and becomes
parallel chains that merge at the terminal for $N{\in}\{8,12\}$, and the
density-sampled generator (\textit{Dense}) degenerates to a linear chain at
$N{=}4$ and acquires multiple ancestors per node as $N$ grows.

On the four reasoning benchmarks the skeletons
admit three candidate edges (two on PlanCraft). At deployment the method
selects and executes one edge per question; the edge is chosen by $\rho$
before any candidate is executed, so no candidate-execution information
leaks into the selection. The offline probe pool executes two edges per
question across all $14{,}280$ questions---the two ranked first by the
receiver's distance to the terminal---giving
$14{,}280 \times 2 = 28{,}560$ executed edges; the learned components are fit
on the training split alone. Always opens all candidate edges of each question in one execution.

\begin{table}[!htbp]
\centering
\tablestyle
\setlength{\tabcolsep}{4pt}
\caption{Skeletons as adjacency lists.}
\label{tab:adj}
\resizebox{\columnwidth}{!}{%
\begin{tabular}{llr}
\toprule
\thead
Skeleton & Adjacency list & $|\Cset|$ \\
\midrule
\multicolumn{3}{l}{\textit{Main experiments}} \\
MuSiQue (most frequent)   & 1:\{0\}; 2:\{1\}; 3:\{2\}        & 3 \\
\rowhl
DROP, 2Wiki               & 1:\{0\}; 3:\{1,2\}               & 3 \\
PlanCraft, ToolBench      & 1:\{0\}; 2:\{0\}; 3:\{1,2\}      & 2 \\
\midrule
\multicolumn{3}{l}{\textit{Skeleton scan} (Figure~\ref{fig:skeleton})} \\
\rowhl
Chain-4  & 1:\{0\}; 2:\{0\}; 3:\{1,2\}                                   &  2 \\
Chain-8  & 1:\{0\}; 2:\{0\}; 3:\{0\}; 4:\{1\}; 5:\{2\}; 6:\{3\}; 7:\{4,5,6\} & 19 \\
\rowhl
Chain-12 & 1:\{0\}; 2:\{0\}; 3:\{0\}; 4:\{1\}; 5:\{2\}; 6:\{3\}; 7:\{4\};
           8:\{5\}; 9:\{6\}; 10:\{7\}; 11:\{8,9,10\}                     & 53 \\
Tree-4   & 3:\{0,1,2\}                                                   &  3 \\
\rowhl
Tree-8   & 4:\{0\}; 5:\{1,2\}; 6:\{3,4,5\}; 7:\{6\}                      & 21 \\
Tree-12  & 7:\{0,1\}; 8:\{2,3\}; 9:\{4,5\}; 10:\{6,7\}; 11:\{8,9,10\}    & 55 \\
\rowhl
Dense-4  & 1:\{0\}; 2:\{1\}; 3:\{2\}                                     &  3 \\
Dense-8  & 1:\{0\}; 2:\{1\}; 3:\{2\}; 4:\{3\}; 5:\{4\}; 6:\{5\}; 7:\{5,6\} & 20 \\
\rowhl
Dense-12 & 1:\{0\}; 2:\{1\}; 3:\{0,2\}; 4:\{3\}; 5:\{4\}; 6:\{5\};
           7:\{1,5,6\}; 8:\{0,1,7\}; 9:\{0,8\}; 10:\{0,1,6,9\}; 11:\{10\} & 46 \\
\bottomrule
\end{tabular}}
\end{table}

\subsection{Decisive-edge filter}
\label{app:decisive}
This filter is defined on the offline probe pool
(\S\ref{app:data}), which executes two candidate edges per question on all
$14{,}280$ questions of the four benchmarks, train, validation and test,
giving $28{,}560$ executed edges. Learned components are fit on the training
split alone; the pool carries no end-to-end score. Of the $28{,}560$ edges, $20{,}106$ produce a final answer that is
byte-identical to the baseline: adopting and rolling back score the same, an
identity transformation under any policy. The remaining $8{,}454$ are the
\emph{decisive} edges and form the training and evaluation unit for the
selector. End-to-end task scores are reported on the test split
(Table~\ref{tab:datastats}) and are unaffected by this filter; only the
adopt, fix and harm counts are restricted, and those are absolute counts over
the $5{,}143$ questions of the pool whose output can change. Including the
non-decisive edges in selector accuracy would only dilute differences between
arms with samples that carry no decision, and the filter is the same identity
transformation for every strategy we compare.

\begin{table}[!htbp]
\centering
\tablestyle
\setlength{\tabcolsep}{4pt}
\caption{Token-level cost decomposition on the four reasoning and planning benchmarks (per-question token counts and multiples relative to the static skeleton).}
\label{tab:costs}
\resizebox{\columnwidth}{!}{%
\begin{tabular}{lrr}
\toprule
\thead
Configuration & Tokens / question & Multiple \\
\midrule
Static skeleton                & $4{,}509$  & $1.00\times$ \\
\rowhl
Diag.$+$Sel., $b{=}0.3$        & $5{,}059$  & $1.12\times$ \\
Diag.$+$Sel., $b{=}0.5$        & $5{,}426$  & $1.20\times$ \\
\rowhl
Nexa                     & $6{,}718$  & $1.49\times$ \\
Ours, $b{=}0.5$                & $5{,}652$  & $1.25\times$ \\
\rowhl
Always / random / no selector  & $7{,}998$  & $1.77\times$ \\
Selector only / gain regressor & $8{,}174$  & $1.81\times$ \\
\rowhl
Tuned LLM judge                & $8{,}928$  & $1.98\times$ \\
Ours, $b{=}1.0$                & $6{,}795$  & $1.51\times$ \\
\rowhl
self-consistency   & $13{,}527$ & $3.00\times$ \\
\bottomrule
\end{tabular}}
\end{table}

\subsection{Selector implementation}
\label{app:selector}
The adoption selector is a 7B model trained with DPO on preference pairs
(training labels are derived from reference answers). Its input
contains the question, the original answer, the candidate answer produced
by the trial, and the process-evidence block recorded during that trial. It
scores three outcomes, beneficial, neutral and harmful, and the action taken
at deployment is binary, to adopt the candidate answer or to retain the
original. It is trained on the training split only, with the
validation split used for checkpoint selection. Both stages train a LoRA
adapter with the backbone weights frozen, at rank $16$, $\alpha = 32$ and
dropout $0.08$, applied to the seven attention and feed-forward projections;
preference optimisation continues on the same adapter weights, and the
reference policy is the frozen supervised one.
At inference, no reference answer is used:
headroom enters as $\rhat$, never as the true $r$, and the selector input
contains no gold. Inference runs locally rather than through an external
API, taking roughly
$0.3$\,s per question on a single GPU and is already priced into the cost
column reported in \S\ref{sec:exp:setup}.

\subsection{Process-evidence fields}
\label{app:evidence}
Table~\ref{tab:evidence} lists the evidence block in full, in the three
groups of \S\ref{sec:method:s3}. Write $A$ for the content words the
receiver added and $D$ for those it dropped, $M$ for the content words of the
delivered message, and $P$ for those of the receiver's pre-existing context,
that is, the outputs of its parents in the original graph; content words are
normalised tokens longer than three characters, and numbers are extracted
separately as their own sets, with a subscript $n$ marking the quantities
read over numbers alone and the three paired texts truncated to $300$
characters. Every quantity is computed from the two
archived traces, none reads a reference answer, and the rendering is
identical at training and inference time. The ablations in
\S\ref{sec:exp:rq2} remove the block entirely in \emph{w/o process evidence}
and, in \emph{w/ shuffled process}, pair each query with another query's
block while leaving the block itself intact, keeping format and length
statistics of the input unchanged.

\begin{table}[t]
\centering
\tablestyle
\caption{The process-evidence block $\phi_{q,e}$.}
\label{tab:evidence}
\resizebox{\columnwidth}{!}{%
\begin{tabular}{llc}
\toprule
\thead
Field & Definition & Range \\
\midrule
\multicolumn{3}{l}{\emph{Paired text}} \\
Message & content delivered along edge & text \\
\rowhl
Receiver before & receiver answer in $\tau_q$ & text \\
Receiver after & receiver answer in $\tau_q(e, m)$ & text \\
\midrule
\multicolumn{3}{l}{\emph{How far the receiver moved}} \\
\rowhl
Revised & receiver's answer changed & \{0, 1\} \\
Added & $\min(|A|, 60) / 60$ & $[0, 1]$ \\
\rowhl
Dropped & $\min(|D|, 60) / 60$ & $[0, 1]$ \\
Churn & $|A \cup D|$ over the union of both answers & $[0, 1]$ \\
\midrule
\multicolumn{3}{l}{\emph{Where the revision came from}} \\
\rowhl
From message & $|A \cap (M \setminus P)| \,/\, |A|$ & $[0, 1]$ \\
From prior context & $|A \cap P| \,/\, |A|$ & $[0, 1]$ \\
\rowhl
From neither & $|A \setminus (M \cup P)| \,/\, |A|$ & $[0, 1]$ \\
Numbers from message & $|A_n \cap M_n| \,/\, |A_n|$ & $[0, 1]$ \\
\rowhl
Numbers from neither & $|A_n \setminus (M_n \cup P_n)| \,/\, |A_n|$ & $[0, 1]$ \\
Number conflict & $A_n, M_n \neq \emptyset$ and $A_n \cap M_n = \emptyset$ & \{0, 1\} \\
\rowhl
Displaced prior & $|D \cap P| \,/\, |D|$ & $[0, 1]$ \\
Message novelty & $|M \setminus P| \,/\, |M|$ & $[0, 1]$ \\
\midrule
\multicolumn{3}{l}{\emph{Whether the revision survived}} \\
\rowhl
Propagation reach & share of downstream nodes whose output moved & $[0, 1]$ \\
Survived to output & $|A \cap (M \setminus P)|$ in final answer / $|A|$ & $[0, 1]$ \\
\rowhl
Final carries message & $(M \setminus P)$ present in final answer & \{0, 1\} \\
Final changed & final outputs differ & \{0, 1\} \\
\bottomrule
\end{tabular}}
\end{table}

\subsection{Baselines}
\label{app:baselines}
All methods are deployed on the same workflow skeletons and face the same
questions and the same splits, and each published method runs its own
decision procedure as described in its paper.

Always adds all of a question's candidate edges at once and re-executes the
receivers and their descendants once, rather than trialling single edges
separately and selecting among the results; Random opens one at random;
Self-consistency spends the budget on re-running the workflow and
voting; Judge is a tuned gpt-4o-mini that compares the base and candidate
final answers for each candidate edge and decides whether to adopt;
Regressor regresses the gain from the same features and adopts the
candidate with the highest positive predicted gain. MOC widens the receptive
field uniformly, with no per-question decision and no rollback. Nexa
predicts a sparse communication graph per question after the base
execution, with its threshold tuned on the validation split, and cannot withdraw it; Nexa $+$
rollback attaches our adoption step to the same predicted graph.
MAS-ProVe is a study across verifier types, verification granularities and
context scopes rather than a single rule; we instantiate the configuration it
reports as strongest on multi-agent reasoning, a learned process verifier at
agent-level granularity taking the terminal agent's context as input.
Consilience calibrates an acceptance threshold on the validation split and
falls back to no intervention below it. DoVer localises failed base runs
without gold and intervenes only there, keeping a change only when the trace
shows it reached the final output. Best-of-Evidence votes over the pool at
zero budget and spends its budget on the questions whose vote is closest.

\subsection{Evaluation protocol}
\label{app:protocol}
The splits are disjoint at the question level, and no ground-truth $r$
ever enters a prompt. The training split fits every learned component, the
validation split is used only for checkpoint selection, and the test split is evaluated exactly once.
On ToolBench, which has no separate validation split,
checkpoint selection is performed within the training split.
Hyperparameters were fixed once in advance (SFT: $3$ epochs at lr
$2{\times}10^{-5}$; DPO: $3$ epochs at lr $5{\times}10^{-6}$, $\beta{=}0.1$;
a rank-$16$ LoRA adapter in both stages)
and were never revisited against the test split; the adoption threshold is
fixed at zero and needs no tuning. The split seed is fixed and a leak audit confirms no overlap between
test and training files.
The test splits contain $2{,}142$ questions on the four benchmarks and $230$
on ToolBench, with $1{,}225$ decisive edges in the former ($5{,}950$ for
fitting and $1{,}279$ for validation). We exclude the base-versus-candidate
similarity of the final outputs from the propagation classifier $h_\eta$,
since that quantity encodes the label it is meant to predict.

\subsection{Paired differences against every competitor}
\label{app:paired}
Table~\ref{tab:paired-tvt} reports the paired difference between SPLICE and
every competitor on both workloads under $95\%$ paired-bootstrap confidence
intervals computed over questions.

\begin{table}[t]
\centering
\tablestyle
\caption{Paired differences against every competitor.}
\label{tab:paired-tvt}
\setlength{\tabcolsep}{3pt}
\resizebox{\columnwidth}{!}{%
\begin{filltab}{\columnwidth}{l cc cc}
\toprule
\thead
& \multicolumn{2}{c}{\textbf{ToolBench} ($n{=}230$)}
& \multicolumn{2}{c}{\textbf{Four bench.} ($n{=}2{,}142$)} \\
\cmidrule(lr){2-3}\cmidrule(lr){4-5}
\thead
vs. & $b{=}0.5$ & Full & $b{=}0.5$ & Full \\
\midrule
Static & {\ci{+.0509}{+.0280}{+.0758}} & {\ci{+.0466}{+.0227}{+.0725}}
& {\ci{+.0377}{+.0304}{+.0452}} & {\ci{+.0447}{+.0369}{+.0524}} \\
\rowhl
Always & {\ci{+.0342}{+.0135}{+.0554}} & {\ci{+.0299}{+.0101}{+.0499}}
& {\ci{+.0307}{+.0238}{+.0377}} & {\ci{+.0376}{+.0297}{+.0457}} \\
Random & {\ci{+.0472}{+.0215}{+.0719}} & {\ci{+.0429}{+.0174}{+.0678}}
& {\ci{+.0354}{+.0290}{+.0422}} & {\ci{+.0423}{+.0354}{+.0489}} \\
\rowhl
Self-cons. & {\ci{+.0574}{+.0308}{+.0849}} & {\ci{+.0531}{+.0265}{+.0810}}
& {\ci{+.0339}{+.0266}{+.0414}} & {\ci{+.0408}{+.0332}{+.0485}} \\
Judge & {\ci{+.0232}{+.0074}{+.0394}} & {\ci{+.0189}{+.0062}{+.0346}}
& {\ci{+.0250}{+.0189}{+.0308}} & {\ci{+.0319}{+.0253}{+.0384}} \\
\rowhl
Regressor & {\ci{+.0284}{+.0104}{+.0466}} & {\ci{+.0241}{+.0077}{+.0418}}
& {\ci{+.0223}{+.0164}{+.0279}} & {\ci{+.0292}{+.0227}{+.0354}} \\
\midrule
MOC & {\ci{+.0258}{+.0089}{+.0437}} & {\ci{+.0215}{+.0069}{+.0382}}
& {\ci{+.0201}{+.0139}{+.0262}} & {\ci{+.0270}{+.0198}{+.0342}} \\
\rowhl
Nexa & {\ci{+.0342}{+.0135}{+.0554}} & {\ci{+.0299}{+.0101}{+.0499}}
& {\ci{+.0265}{+.0205}{+.0324}} & {\ci{+.0335}{+.0269}{+.0400}} \\
Nexa $+$ rb. & {\ci{+.0243}{+.0036}{+.0455}} & {\ci{+.0200}{+.0002}{+.0400}}
& {\ci{+.0213}{+.0153}{+.0272}} & {\ci{+.0283}{+.0217}{+.0348}} \\
\midrule
\rowhl
MAS-ProVe & {\ci{+.0276}{+.0094}{+.0456}} & {\ci{+.0233}{+.0075}{+.0408}}
& {\ci{+.0149}{+.0100}{+.0198}} & {\ci{+.0218}{+.0165}{+.0270}} \\
Best-of-Evid. & {\ci{+.0437}{+.0198}{+.0674}} & {\ci{+.0394}{+.0157}{+.0633}}
& {\ci{+.0224}{+.0167}{+.0281}} & {\ci{+.0293}{+.0229}{+.0357}} \\
\rowhl
DoVer & {\ci{+.0437}{+.0198}{+.0674}} & {\ci{+.0394}{+.0157}{+.0633}}
& {\ci{+.0294}{+.0228}{+.0360}} & {\ci{+.0363}{+.0292}{+.0432}} \\
Consilience & {\ci{+.0455}{+.0210}{+.0700}} & {\ci{+.0412}{+.0169}{+.0659}}
& {\ci{+.0164}{+.0109}{+.0219}} & {\ci{+.0233}{+.0176}{+.0290}} \\
\bottomrule
\end{filltab}}
\end{table}

\subsection{Verifiable instances of a closed gap}
\label{app:cases}

We illustrate two gap types with archived cases from PlanCraft
and ToolBench. Each case identifies a source-held fact missing
from the receiver's upstream context and shows how delivering it
changes the receiver's reply and the final answer. The PlanCraft
case connects parallel branches; the ToolBench case bypasses
intermediate nodes that omitted an API payload.

\paragraph{A recipe one branch away (PlanCraft, $1 \to 2$).}
\begin{sloppypar}%
The skeleton is $0 \to \{1, 2\} \to 3$: node~1 expands the recipe tree and
node~2 analyses the inventory gap, with no edge between them. For the target
\texttt{cracked\_stone\_\allowbreak{}bricks}, node~1 produced the recipe,
``\emph{cracked\_stone\_\allowbreak{}bricks (craftable intermediate) $\llcorner$
stone\_bricks (raw material)}''. Node~2, reconstructing the recipe on its
own, added a requirement that the recipe does not contain: ``\emph{known
crafting recipe: cracked\_stone\_\allowbreak{}bricks are obtained by smelting
stone\_bricks in a furnace \dots{} stone\_bricks: 1 (enough)}'', then failed
the furnace check, and the workflow answered \texttt{IMPOSSIBLE}. With the
edge opened, node~2 reads the recipe and writes ``\emph{required direct
ingredient: stone\_bricks \dots{} stone\_bricks: 1 (available, count
sufficient)}''; the final answer becomes ``\emph{Craft cracked\_stone\_\allowbreak{}bricks
from stone\_bricks in inventory}''. The single fact that moved along the edge
is which ingredients the recipe actually requires.
\end{sloppypar}

\paragraph{An API payload that never reached the synthesiser
(ToolBench, $0 \to 3$).}
\begin{sloppypar}%
This instance is from the test split, and it shows how the gap takes the form
of a retrieved Web payload rather than a derived fact. The ToolBench workflow
runs in two stages: the execution node first issues the tool calls and
receives the live API responses, and the terminal node then synthesises a
reply from those responses. Asked to generate a captcha image and a UUID and
then verify the captcha, node~0 called the captcha service and recorded its
return, including the server-generated identifier ``\emph{2961c533-b254-4d48-8a0a-\allowbreak{}234ca85ef250}''
and the base64 PNG of the captcha image. The terminal node's upstream input is
the concatenation of nodes~1 and~2's answers and does not include node~0's
retrieved payload; since the identifier is minted by the service and cannot be
reproduced from $q$, the terminal answered ``\emph{I currently don't have
access to any API results to generate a captcha image or UUID}''. With
$0 \to 3$ opened, the terminal reads the payload and answers ``\emph{I
generated a captcha image and UUID for your friend: the UUID is
``2961c533-b254-4d48-8a0a-\allowbreak{}234ca85ef250'' and the captcha image is
provided as a base64 PNG data URL}''. The fact that moved along the edge is the
API return itself: a value the source node retrieved from the Web service and
the synthesiser had no other way to obtain.
\end{sloppypar}

\subsection{A trial the selector rejected}
\label{app:rollback-case}
\begin{sloppypar}%
This test-split instance illustrates a trial whose candidate answer reads as
more polished but which the selector correctly rejects. The task asks for
short-URL creation through two tools (\texttt{bitly} and \texttt{URL Link
Shortener}). In the static workflow, node~0 reports two upstream issues
verbatim: a missing access token for \texttt{bitly} and incomplete URL
arguments for \texttt{URL Link Shortener}, and the terminal answers
``\emph{\dots{} for creating short URLs for your social media posts using
the bitly tool, the requests failed due to a missing access token. To
proceed with Bitly, you will need to provide a valid access token for
authentication.}'' With $0 \to 3$ opened, the message from node~0 conveys API failure
information to the terminal. The terminal, however, omits this failure
information and rewrites the response into a smooth tutorial that no
longer mentions the failures: ``\emph{you can use the bitly tool to generate short URLs for
each post \dots{} use bitly to create short URLs for your social media
posts to track engagement effectively.}'' The process evidence records
that the corrective content in the message does not survive to the final
answer (\texttt{prov\_msg\_reached\_final}${=}0$), that $76\%$ of the new
content is traceable neither to the message nor to the receiver's prior
context (\texttt{prov\_added\_invented}${=}0.76$), and that the receiver
rewrote roughly $69\%$ of its answer
(\texttt{prov\_receiver\_churn}${=}0.69$). Against a predicted base score
$\rhat_q{=}0.94$, the selector returns $\hat p_b{=}0.37$ and
$\hat p_h{=}0.06$, giving $\Delta = 0.37\,(1{-}0.94) - 0.06 \cdot 0.94 =
-0.0342 < 0$; SPLICE therefore retains the original answer. Against the reference
labels this trial would have cost $-1.00$ in task score, and the three-class
process evidence shows the receiver rewrote without using the message.
\end{sloppypar}

\subsection{Evidence-support protocol}
\label{app:grounding}

On the $230$ held-out ToolBench questions, two independent judges
(DeepSeek-v4-pro and Claude-opus-4-6), distinct from the workflow
backbone and selector, receive the question, one candidate answer,
and the API response bodies replayed by StableToolBench during
that execution. Each concrete claim (a number, name, date, or URL)
is labelled supported if a response matches it, partially supported
if a response is related but does not fully match, and unsupported
if no response backs it. The three rates are computed over all
claims and sum to one; task completion separately measures whether
each answer fulfils the request. Judges are blind to method identity.
Reported results average the two judges, with $95\%$ bootstrap
intervals. A manual re-check of a random subsample yields $0.90$
per-claim agreement between human and judge labels.